\documentclass[conference]{IEEEtran}
\IEEEoverridecommandlockouts
\usepackage{cite}
\usepackage{amsmath,amssymb,amsfonts}
\usepackage{algorithmic}
\usepackage{graphicx}
\usepackage{textcomp}
\usepackage{xcolor}
\usepackage{lipsum}
\usepackage{url}

\def\BibTeX{{\rm B\kern-.05em{\sc i\kern-.025em b}\kern-.08em
    T\kern-.1667em\lower.7ex\hbox{E}\kern-.125emX}}
\begin{document}

\title{MetaPix: A Data-Centric AI Development Platform for Efficient Management and Utilization of Unstructured Computer Vision Data\\
}

\author{\IEEEauthorblockN{Sai Vishwanath Venkatesh}
\IEEEauthorblockA{\textit{Ford Motor Company} \\
Dearborn, Michigan \\
svenk193@ford.com}
\and
\IEEEauthorblockN{Atra Akandeh}
\IEEEauthorblockA{\textit{Ford Motor Company} \\
Dearborn, Michigan \\
aakandeh@ford.com}
\and
\IEEEauthorblockN{Madhu Lokanath}
\IEEEauthorblockA{\textit{Ford Motor Company} \\
Dearborn, Michigan \\
mlokanat@ford.com}
}

\maketitle

\begin{abstract}
In today's world of advanced AI technologies, data management is a critical component of any AI/ML solution.  Effective data management is vital for the creation and maintenance of high-quality, diverse datasets, which significantly enhance predictive capabilities and lead to smarter business solutions. In this work, we introduce MetaPix, a Data-centric AI platform offering comprehensive data management solutions specifically designed for unstructured data. MetaPix offers robust tools for data ingestion, processing, storage, versioning, governance, and discovery. The platform operates on four key concepts: DataSources, Datasets, Extensions and Extractors. A DataSource serves as MetaPix top level asset, representing a narrow-scoped source of data for a specific use. Datasets are MetaPix second level object, structured collections of data. Extractors are internal tools integrated into MetaPix's backend processing, facilitate data processing and enhancement. Additionally, MetaPix supports extensions, enabling integration with external third-party tools to enhance platform functionality. This paper delves into each MetaPix concept in detail, illustrating how they collectively contribute to the platform's objectives. By providing a comprehensive solution for managing and utilizing unstructured computer vision data, MetaPix equips organizations with a powerful toolset to develop AI applications effectively.
\end{abstract}

\section{Introduction}
In the modern business landscape, data has emerged as one of the most valuable assets for companies. The rapid advancement of artificial intelligence (AI) and machine learning (ML) technologies has further elevated the importance of data, transforming it into a strategic asset that drives innovation, competitive advantage, and growth. Machine learning models rely on high-quality, diverse, and well-managed data as a solid foundation. Therefore, organizations must implement a comprehensive data management strategy to gain competitive advantages.

Effective data management is essential for any AI/ML solution. It ensures that data used for training and validation is accurate, consistent, and accessible. Data management involves several critical aspects, including data ingestion, data processing and enhancement, data storage, data versioning, data governance, and data discovery. Data ingestion involves collecting data from diverse sources, including databases, APIs, and streaming platforms, using both batch and real-time methods. Data processing and enhancement include cleaning and preparing raw data for model training. Data storage solutions, such as data lakes, data warehouses, and databases, provide organized and scalable storage. Data versioning tracks changes and maintains different dataset versions, enhancing reproducibility. Data governance ensures access control, data integrity, and compliance. Data cataloging enables dataset discovery and management by providing detailed metadata and search capabilities. These integrated processes enable efficient and compliant data management, which forms the backbone of a robust AI/ML solution.

This work presents MetaPix, a Data-centric AI platform that provides comprehensive data management solutions tailored for unstructured data. MetaPix consists of four key components: DataSource, Dataset, Extractors, and Extensions. These components collectively ensure the provision of high-quality and reliable data for machine learning workflows. In MetaPix, DataSources manage data ingestion, storage, and governance, while Datasets addresses data versioning. Extractors and Extensions also facilitate data processing, enhancement, search, and discovery, ensuring a robust and efficient data management ecosystem. The final product of MetaPix is a connected golden dataset, which is a high-quality, well-curated dataset that serves as a benchmark for training and evaluating machine learning models. This golden dataset results from comprehensive data management practices that ensure data quality, integration, consistency, and governance.

This paper provides an in-depth examination of each MetaPix concept, highlighting its significance in achieving the platform’s overarching objectives. The following sections are structured as follows: we begin with an overview of two related work, followed by a detailed exploration of each core component.

\section{Related Work}
\cite{ismail2019premise} from MIMOS research institute outlined their blueprint to build and host an internal on-premise AI platform. Leveraging their existing infrastructure, including private cloud services, distributed storage, and a unified authentication platform, they introduced key components of their platform such as Mi-Cloud, Mi-ROSS, Mi-Focus, and Mi-Registry. VM disks are hosted on top of Mi-ROSS distributed storage, while AI training and inferencing are facilitated through containerization on their private cloud infrastructure. Container management is handled by Mi-Focus, built on Kubernetes, with Mi-Focus Registry managing application versioning and AI model storage. Their enhancements to the Docker registry include distribution to boost performance during high loads. Additionally, they integrated vulnerability assessment tools to aid in pre-deployment troubleshooting. Deployment options include Mi-Focus Container for cloud deployments and Mi-Focus Edge for edge computing scenarios. This setup empowers their users like data and ML engineers to seamlessly create datasets on their laptops and mount them into containers for ML training and validation processes.

A 2023 paper \cite{mazumder2024dataperf} presented at Neural NIPS introduces the DataPerf competition, which focuses on evaluating dataset creation and quality to advocate for data-centric approaches over traditional model-centric evaluation methods. Through this initiative, the authors explore methods to benchmark data-centric pipelines and establish best practices for these approaches. The paper's contributors, from the MLCommons group, include both academic researchers and industry practitioners.

\section{DataSources}
\label{DatasourceSection}

DataSources are being used to manage data ingestion, storage, and governance. A DataSource is created by a team on the platform that identifies a stream of data that could be high in volume, velocity, and reuse. This functions as queryable live data enabled with Google Big Query and used downstream to create datasets. The authors define “DataSource” as a data abstraction for production data that serves as a queryable superset.


\begin{figure}[!htb]
	\centering
	\includegraphics[trim={0 0 0 0},clip,scale=0.22]{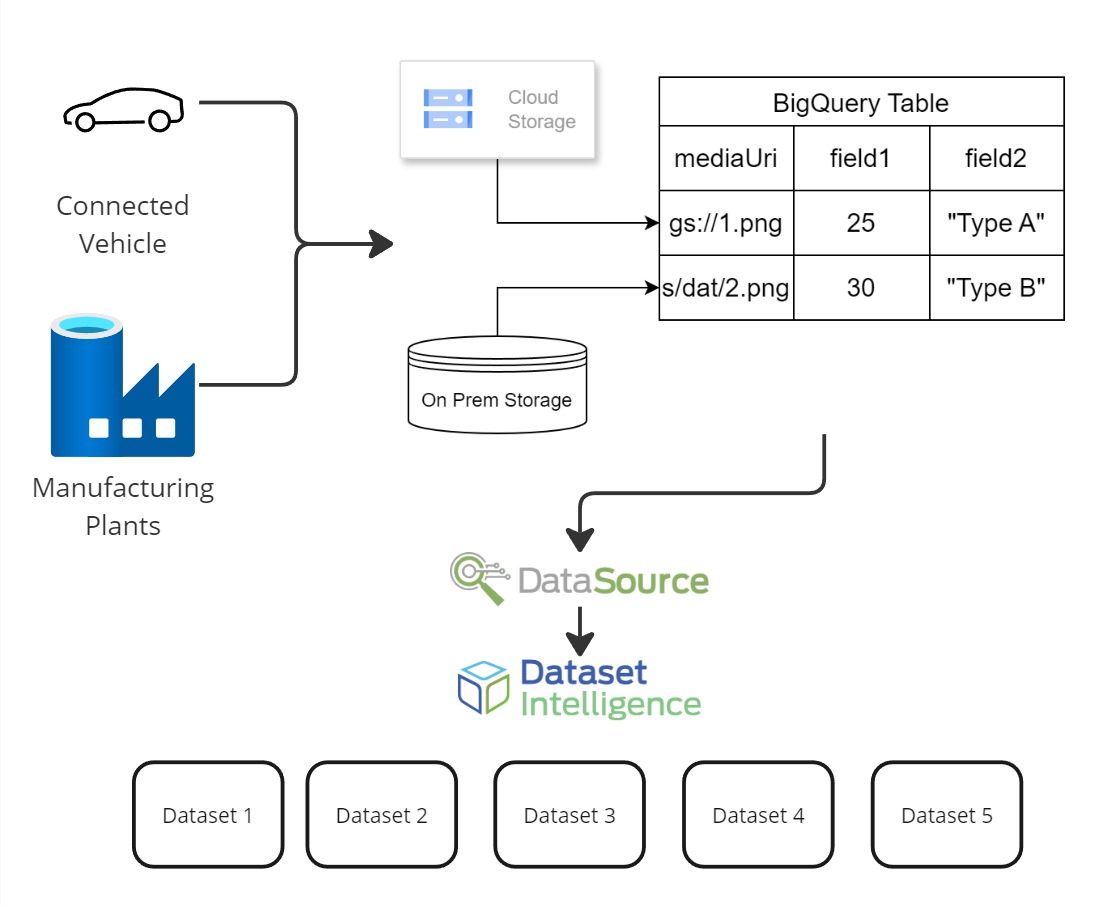}
	\caption{MetaPix Datasource Architecture}
	\label{MetaPixDatasourceArch}
\end{figure} 

Computer Vision and unstructured data, originating daily from connected vehicles or manufacturing plants, are stored either on-premises or in GCP Cloud Storage Buckets. These buckets are connected to a GCP object table, which indexes all media using a unique generation ID. When new images or media are added to these storage buckets, the corresponding index changes are pushed to the object table. This object table is linked to a BigQuery table, which contains additional metadata such as vehicle type and location, via the media generation ID. This combined table is referred to as the extended attribute table. This setup is depicted in Figure \ref{MetaPixDatasourceArch}. MetaPix has developed data crawlers to monitor and search media stored on on-premises servers to populate these tables.  A view created from merging these two tables is then associated with a DataSource, facilitating user access to this data. 



In addition to linking this BigQuery table to a DataSource, we specify several properties to enhance data governance and lineage. This framework allows developers and teams to securely create their datasets using a DataSource, as detailed in Table \ref{DatasourceObject}. A designated 'Data Owner'—usually an organizational leader or an expert familiar with the data—manages the DataSource's creation and ongoing maintenance. The Data Owner collaborates with the Office of General Counsel (OGC) to evaluate data and privacy risks, as reflected by the property 'Cat\_level'. This process also identifies the relevant organization for the data and establishes the roles or security groups permitted restricted access.


When setting up a DataSource, the Data Owner supplies essential information to enable the backend crawling service to crawl or monitor new data. This information includes the GCP Project ID, table name, GCP BigQuery view for index creation, and the column name that holds the target media location (mediaUri). For dynamic data, as opposed to static, the Data Owner may also specify a list of storage locations for the crawling service to monitor for new files. These locations are recorded in the 'storage\_locations' property.

Considering the DataSource serves as a comprehensive data superset before teams branch out to address specific AI/ML challenges by creating datasets, the Data Owner has the option to include a visualization link. This link assists teams in understanding the data prior to dataset creation. Furthermore, additional properties like 'region', 'media\_count', and 'last\_modified\_date' are included to provide a more detailed description of the data.

The DataSource features a 'storage\_system' property that specifies whether the data is hosted on GCP or on-premises. This detail helps data scientists plan their compute provisioning in advance of dataset creation. Once a name and description are assigned, along with the specified properties, each DataSource receives a unique DataSource\_id. This ID is crucial for tracking all datasets derived from it. Having detailed information about the root-level source data simplifies the process of implementing changes across all datasets originating from the DataSource.



Another significant efficiency gained at the root level is the generation of vector embeddings for images during the creation of the DataSource. Since generating embeddings is typically a time-consuming process, creating them during the DataSource setup allows all subsequent datasets derived from it to immediately leverage these embeddings. This setup also enables the creation of datasets from text-based searches, enhancing efficiency and functionality.


\begin{table}[h]
\caption{Datasource Object Description}
\centering
\begin{tabular}{|l|l|}
\hline
\multicolumn{1}{|c|}{\textbf{Key}} & \multicolumn{1}{c|}{\textbf{Description}}                             \\ \hline
\_id                      & Datasource unique id                                         \\ \hline
name                      & Datasource name                                             \\ \hline
description               & Datasource description                                      \\ \hline
Security category level   & An integer that classifies data privacy risk                 \\ \hline
project\_id               & Google Cloud Project ID for Data                             \\ \hline
dataset\_id               & Google Cloud Dataset name inside project                     \\ \hline
view                      & Reference BigQuery Table View name                           \\ \hline
media\_uri\_field         & Target URIs                \\ \hline
storage\_locations        & List of GCS bucket names referenced in table                 \\ \hline
visualization\_link       & Link to exploratory visualization for corpus                 \\ \hline
region                    & List of regions where the data is sourced                    \\ \hline
data\_owner               & Data owner email ID - To escalate for access                 \\ \hline
organization              & Organization to which the data belongs                       \\ \hline
storage\_system           & GCP or On-Prem Storage                                       \\ \hline
access\_level             & Unrestricted/Public or Gated Access                          \\ \hline
roles                     & List of user groups with data access                         \\ \hline
media\_count              & Number of records                                            \\ \hline
last\_modified            & Last modified datetime                                       \\ \hline
create\_date              & Created datetime                                             \\ \hline
\end{tabular}
\label{DatasourceObject}
\end{table}

As demonstrated, Datasources on the MetaPix platform are essential for effective data management and utilization, offering a secure and streamlined approach to accessing, managing, and preparing unstructured computer vision data for AI development.

\section{Datasets}
\label{DatasetSection}




A MetaPix dataset is a second-level abstraction within the MetaPix platform. It can either be a subset or a derivative of a DataSource, or it may represent a complete dataset on its own. Importantly, this dataset is logical, storing only metadata and paths to the media, not the media itself. This design enables users to version and link their data to AI-powered tools and services on the platform. The MetaPix Dataset Intelligence acts as a workbench, supporting the exploration of datasets and their integration with connected services. It allows scientists to iteratively refine a dataset specifically tailored to a particular use case by enhancing, enriching, and annotating the data. Key features of MetaPix’s Dataset Intelligence include data quality monitoring, versioning, data lineage, and preventing data duplication. Dataset Intelligence achieves significant storage cost savings by efficiently managing data duplication—referencing media files multiple times, with 500,000 references to 150,000 physical files.


Teams with data already stored in a MetaPix DataSource can leverage the MetaPix Dataset to explore and advance their projects. Multiple teams can concurrently use the same data, each working with distinct versions. This flexibility is particularly beneficial for teams testing proof of concepts before transitioning to production. Using Dataset Intelligence, they can create and manage multiple versions of logical datasets, enabling thorough exploration and refinement. This process aids in developing a more effective dataset, optimized for production readiness. To create a dataset object, users must provide a prepared JSONL or COCO JSON file specifying GCS storage locations or on-premise locations to reference the file path. Alternatively, they can execute an SQL query on a Datasource. Table \ref{DatasetObject} details the properties of a dataset collection developed through Dataset Intelligence. 

A critical property, the 'storage\_system' parameter, informs both the backend service and the user of the data's storage location, whether on-cloud or on-premises. This distinction enables Dataset Intelligence to more efficiently allocate compute and other resources, optimizing proximity and responsiveness to the data. For instance, on-premises datasets can benefit from faster and more economical GPU resources, whereas GCP-stored datasets enjoy faster access to GCP-specific services. The 'embeddings\_enabled' parameter is an optional setting that users can select during the creation of a dataset, indicating if they wish to have CLIP-based vector embeddings generated immediately for their images or videos. These embeddings are crucial for enhancing Data Discovery and Search functionalities, as detailed in section \ref{Extractors}. If a dataset is derived from a DataSource that already possesses these embeddings, users do not need to wait for new embeddings to be generated, thus allowing immediate access to features like search, visualization.

The "visibility" property controls roles and access, ensuring that when a dataset is derived from a DataSource, the same access roles and groups apply to the dataset. This functionality streamlines the sharing process with team members, enabling quick collaboration among those who already have access while upholding data governance standards. Furthermore, the "license" property offers data scientists guidelines on the fair use of the data for downstream applications, ensuring compliance and ethical use. The "data\_source" property is an optional parameter that captures the parent DataSource's information, as referenced in Table \ref{DatasourceObject}. This includes specifics about the DataSource object to support data lineage, access control, and organization of datasets by DataSource. Grouping datasets in this way allows teams to easily access existing work, thereby facilitating rapid innovation within their specific domain. Each dataset is linked to a unique "\_id," serving as the primary key for dataset queries.

The "versions" property records the development of a dataset through its progressive stages of enhancement and refinement. For instance, in an specific use case for creating an annotated computer vision dataset from unlabeled dashcam footage, the initial version "v1" might mask faces and number plates to protect personally identifiable information (PII), while "v2" could add annotations for lanes and cars. A later version might integrate detailed ground truth annotations provided by external annotators using an annotation tool. This property helps track each step in the dataset's evolution, ensuring a systematic and traceable enhancement process. Besides tracking versions, Dataset Intelligence also oversees the use of AI applications, extractors (as discussed in section \ref{Extractors}), and extensions (noted in section \ref{extensions}) applied to the data. This monitoring provides users with insights into the dataset's impact, detailing how the data has been utilized and shared, and informing continuous improvement and collaboration.

\begin{table}[h]
\caption{Dataset Object Description}
\centering
\begin{tabular}{|l|l|}
\hline
\multicolumn{1}{|c|}{\textbf{Key}} & \multicolumn{1}{c|}{\textbf{Description}}                 \\ \hline
\_id                      & Dataset unique id                                \\ \hline
cdsid                     & Dataset creator                                  \\ \hline
name                      & Dataset Name                                     \\ \hline
description               & Dataset Description                              \\ \hline
tags                      & Ddataset tags to aid search                      \\ \hline
license                   & Licensces associated with the dataset            \\ \hline
versions                  & List {[}{]} Containing metadata for all versions \\ \hline
Datasource                & Dictionary with Datasource info if applicable    \\ \hline
visibility                & "Public" or "Restricted"                         \\ \hline
storage\_system           & "GCP" or "On -Prem"                              \\ \hline
embeddings\_enabled       & Indicates if embeddings are linked with the data \\ \hline
has\_annotations          & Indicates if data has external annotations       \\ \hline
\end{tabular}
\label{DatasetObject}
\end{table}






\section{Extractors}
\label{Extractors}

MetaPix has developed several in-house solutions for data processing and enhancement. Extractors are AI-powered tools integrated into MetaPix's backend processing pipeline. Utilizing advanced computer vision and natural language processing techniques, MetaPix's extractors automatically extract meaningful information from unstructured data. Some implemented solutions include Monocular Depth Estimation, Personally Identifiable Information Blur, and Image Segmentation. This section focuses on one of MetaPix's primary features: embedding-based search, which enables customers to perform semantic search and data discovery.

MetaPix provides users with an effective method to discover new data through embedding-based search. This tool enables users to find data relevant to their use cases using semantic search, eliminating the need for user-generated descriptions or tags. Additionally, users can create new datasets directly from the search results (refer to section \ref{DatasourceSection}). The MetaPix search functionality is accessible via the MetaPix platform, MetaPix API and MetaPix Python client SDK. To manage costs and resources effectively, MetaPix leverages on-premises resources to handle GPU-intensive embedding creation. Other services used in this process include GCP Pub/Sub, MongoDB, and Elasticsearch. Pub/Sub is an asynchronous and scalable messaging service that decouples services producing messages from services processing those messages. Elasticsearch functions as a vector database, while MongoDB serves as the primary database. Below is a detailed description of the MetaPix embedding-based search service.


Embedding-based search involves two key processes. The first process generates an embedding for the content added, such as an image or video, and stores the resulting vector in a vector database. The second process generates an embedding for the query or search string which is used to retrieve the most similar vectors from the indexed dataset. These embeddings are created using CLIP (Contrastive Language-Image Pretraining), a deep learning model developed by OpenAI in 2021. CLIP is designed to create embeddings for images/videos and text within a shared space, enabling direct comparisons between the two modalities. This is accomplished by training the model to bring related images/videos and texts closer together in the embedding space while distancing unrelated pairs.

\begin{figure}[!htb]
	\centering
	\includegraphics[trim={0 0 0 0},clip,scale=0.42]{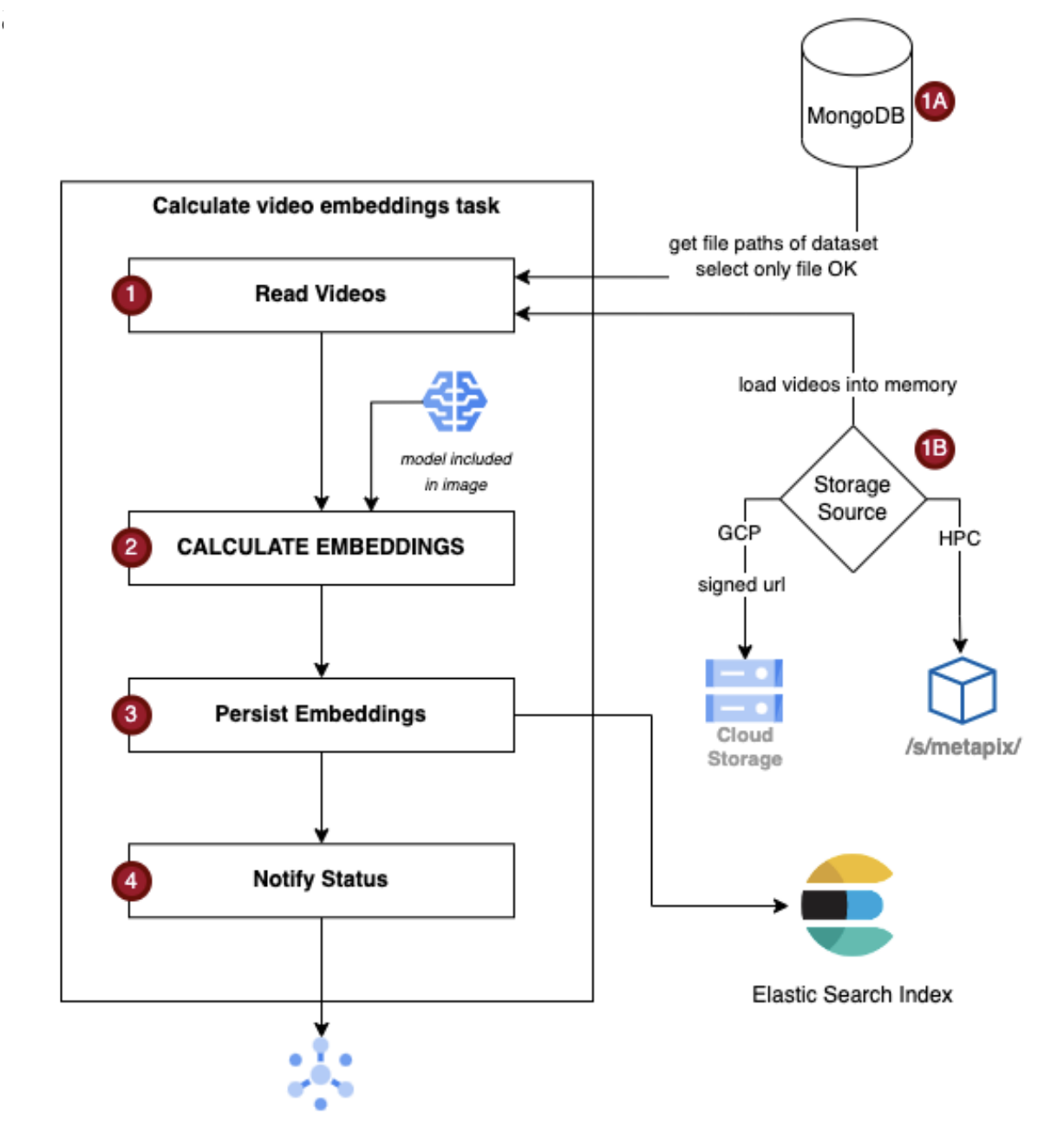}
	\caption{Schematic Representation of the Embeddings Creation Process}
	\label{EmbeddingsCreation}
\end{figure}
 
The first process begins when a user intends to create a dataset through the MetaPix UI (see section \ref{DatasetSection}). Once the files are uploaded and stored in either cloud or on-premises storage, a Pub/Sub message initiates the embedding creation process. MongoDB is then updated with the operation ID for calculating video embeddings. Subsequently, the UI service sends a request to the MetaPix search service which submits a batch job with instructions to run the necessary script. This batch job is executed in a Kubernetes container. Once the embeddings are calculated and stored in Elasticsearch, with KNN search enabled on the index, they become accessible for consultation through similarity search when a query is presented.

\begin{figure}[!htb]
	\centering
	\includegraphics[trim={0 0 0 0},clip,scale=0.28]{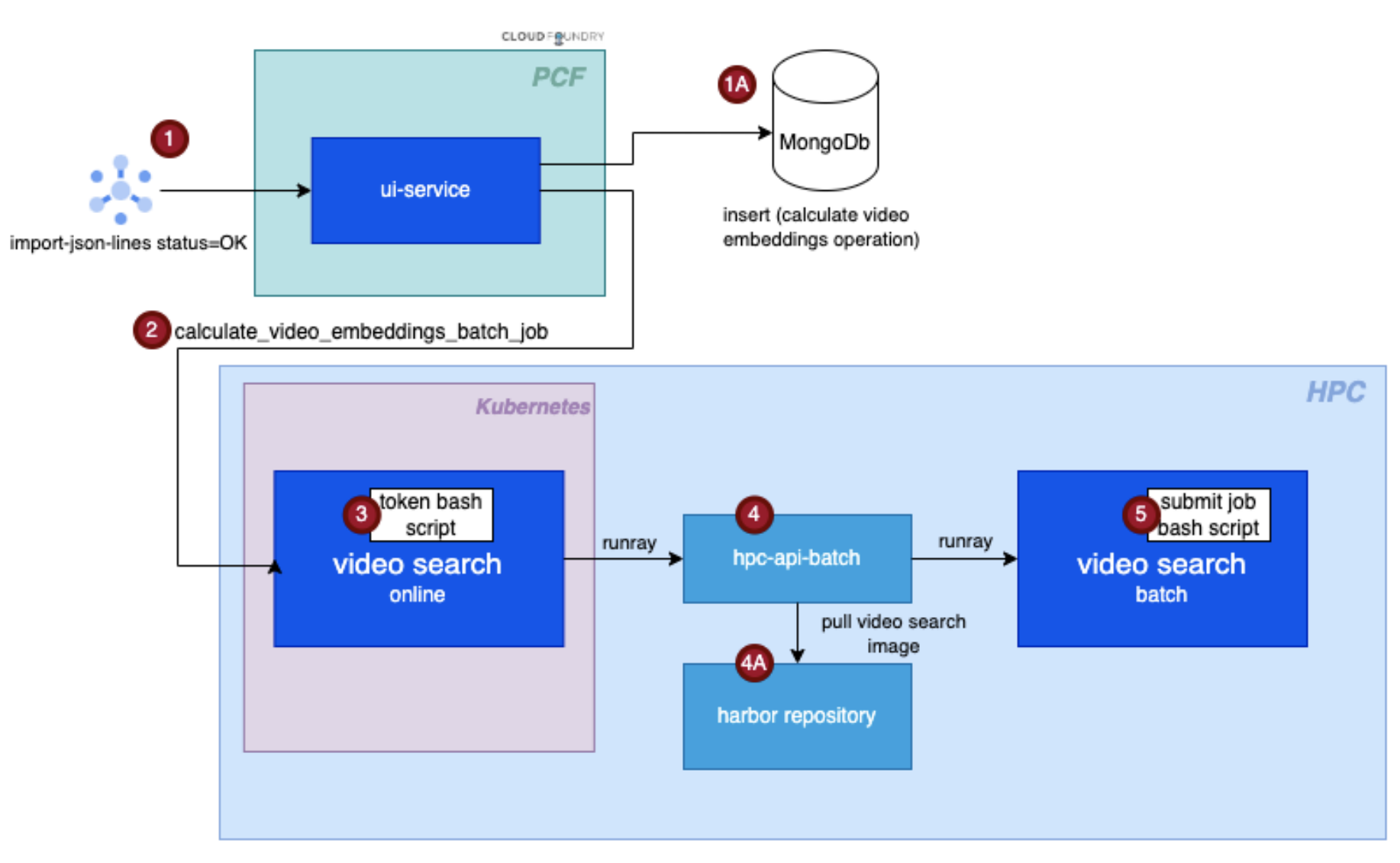}
	\caption{Workflow Diagram of Batch Job Submission to Calculate Embeddings}
	\label{BatchJob}
\end{figure}
 
In the second process, when a user types a query into the search box on the front end (MetaPix UI), the backend service sends a request to the MetaPix search service. The service then generates a vector from the text using the same model employed to calculate image/video embeddings. This query vector is then used to retrieve relevant segments through an approximate K-nearest neighbors search. Finally, The MongoDB collection is consulted using a dataset ID and version and a list of relevant segments is returned to the UI for visualization. The search results are returned in smaller, consumable segments, enhancing usability and data accessibility.


\section{Extensions}
\label{extensions}
In the fast-paced realm of AI, building a data-centric platform is time-intensive. Collaborating with vendors who specialize in developing features that would otherwise take too long to create is crucial. This is why partnerships with vendors are essential to enhance MetaPix with the most effective tools available. Extensions include any fully-fledged tools that lie outside the MetaPix Ecosystem, such as an annotation studio, a data visualization platform, or a model tracking tool. These extensions integrate seamlessly to enrich the capabilities of MetaPix, ensuring it remains at the forefront of technological advancement.

The Annotations service facilitates the integration of extensions by ensuring that each external tool adheres to a common format for interacting with MetaPix Datasets. By defining data formats and establishing Parsers, MetaPix can connect seamlessly with various use cases and platforms. For instance, a user can link a MetaPix Dataset to an Annotation Studio extension by exporting the data in a format compatible with the tool. Once annotated, the dataset, enriched with the newly annotated data, can be re-imported into MetaPix using a Parser that supports the annotation type. This streamlines the workflow and enhances the dataset's utility across different applications.

\begin{figure}[!htb]
	\centering
	\includegraphics[trim={0 0 0 0},clip,scale=0.36]{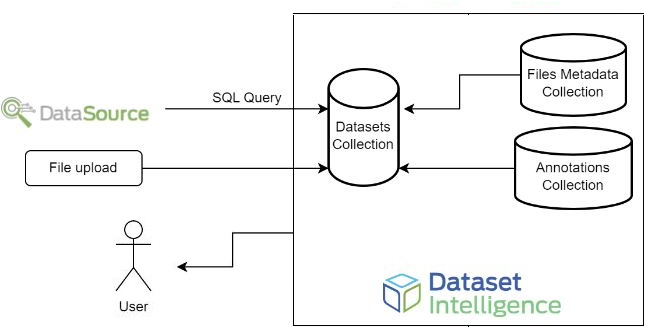}
	\caption{Storage Architecture for Datasets Metadata}
	\label{DatasetsMetadataStorage}
\end{figure}

MetaPix Annotations comprise an additional collection of files designed to store metadata related to external annotations. This capability is essential for processing and supporting various dataset types within a platform that handles unstructured data. By storing annotations and their associated metadata, MetaPix can utilize defined Parsers and Exporters to facilitate the import and export of unstructured data to other tools and use cases within the platform. Figure \ref{DatasetsMetadataStorage} depicts how the annotations collections and files collection are linked to the dataset object collection. Each dataset and version has a list of linked annotations. Each annotation object in this list is detailed with properties outlined in Table \ref{annotationObjectTable}. This metadata is crucial for the MetaPix backend service, enabling users to connect a version of their dataset for import and visualization on external platforms, such as a dedicated annotation platform.

The "type" value in an annotation specifies its category, facilitating the parsing of the "properties" variable content. The "properties" variable is a dictionary containing type-specific attributes. For example, a COCO-formatted annotation \cite{lin2014microsoft} includes properties such as "COCO\_file\_path" and "root\_dir." Similary, for a BigQuery annotation type, it stores attributes like "query\_used" and "Datasource\_info". These are utilized by COCO and BigQuery Parsers, respectively, to access the source file or to import annotations into a new context. Each annotation object in the list features a unique identifier "\_id," a user-defined name, and links to both the dataset and its specific version. Supported by a dedicated annotations service, various data formats are accommodated, facilitating multimodal viewing. Formats such as OpenLabel \cite{asamOpenLabel} and COCO Format, which are ideal for object detection, are easily parsable and linked directly to the corresponding MetaPix dataset ID and version. This setup significantly enhances data handling and integration capabilities.

\begin{table}[h]
\caption{Annotation Object Description}
\begin{tabular}{|l|l|}
\hline
\multicolumn{1}{|c|}{\textbf{Key}} & \multicolumn{1}{c|}{\textbf{Description}}            \\ \hline
\_id                               & Unique annotation id                                 \\ \hline
dataset\_id                        & dataset\_id linked to annotation                     \\ \hline
version                            & version from dataset linked to annotation            \\ \hline
name                               & User provided name for annotation file               \\ \hline
is\_default                        & If this is the deafault metadata view for this data. \\ \hline
type                               & Annotation type - COCO, YOLO, SQL Query etc          \\ \hline
properties                         & Dictionary with type specific properties             \\ \hline
create\_date                       & Created date                                         \\ \hline
\end{tabular}
\label{annotationObjectTable}
\end{table}


\section{Conclusion}
Effective data management is vital for any organization aiming to harness its data assets for strategic gains. In this work we outlined a comprehensive data management solution tailored specifically for unstructured data. We addressed the implementation of essential components of a successful data management system, including data ingestion, processing and enhancement, storage, versioning, governance, and discovery. By prioritizing data quality and accessibility, organizations can enhance decision-making, boost operational efficiency, adhere to regulatory standards, and ultimately fulfill their business objectives.

\bibliographystyle{plain}
\bibliography{ref}

\end{document}